\documentclass[%
 aip,
 amsmath,amssymb,
preprint,
]{revtex4-2}

\usepackage{graphicx}
\usepackage{dcolumn}
\usepackage{bm}

\usepackage[utf8]{inputenc}
\usepackage[T1]{fontenc}
\usepackage{mathptmx}
\usepackage{etoolbox}
\usepackage{caption}
\usepackage{subcaption}
\usepackage{multirow}
\usepackage[ruled]{algorithm2e}

\makeatletter
\def\@email#1#2{%
 \endgroup
 \patchcmd{\titleblock@produce}
  {\frontmatter@RRAPformat}
  {\frontmatter@RRAPformat{\produce@RRAP{*#1\href{mailto:#2}{#2}}}\frontmatter@RRAPformat}
  {}{}
}%
\makeatother
\begin{document}


\title[]{Multi-level datasets training method in Physics-Informed Neural Networks}
\author{Yao-Hsuan Tsai$^1$}
\author{Hsiao-Tung Juan$^1$}
\author{Pao-Hsiung Chiu$^{2,*}$}
\author{Chao-An Lin$^{1,*}$}
\affiliation{ 
$^1$Department of Power Mechanical Engineering, National Tsing Hua University, Hsinchu 30013, Taiwan\\
$^2$Institute of High Performance Computing, Agency for Science, Technology and Research (A*STAR), Singapore\\
$^*$\footnotesize corresponding authors. Email: calin@pme.nthu.edu.tw(C.A. Lin); chiuph@ihpc.a-star.edu.sg(P.H. Chiu)
}%
\keywords{Physics-informed neural networks, Multi-level datasets training, Dimensional analysis weighting}



\begin{abstract}
Physics-Informed Neural Networks (PINNs) have emerged as a promising methodology for solving partial differential equations (PDEs), gaining significant attention in computer science and various physics-related fields. Despite being demonstrated the ability to incorporate the physics of laws for versatile applications, PINNs still struggle with the challenging problems which are stiff to be solved and/or have high-frequency components in the solutions, resulting in accuracy and convergence issues. It will not only increase computational costs, but also lead to accuracy loss or solution divergence, in the worst scenario. In this study, an alternative approach is proposed to mitigate the above-mentioned problems. Inspired by the multi-grid method in CFD community, the underlying idea of the current approach is to efficiently remove different frequency errors via training with different levels of training samples, resulting in a simpler way to improve the training accuracy without spending time in fine-tuning of neural network structures, loss weights as well as hyper-parameters. To demonstrate the efficacy of current approach, we first investigate the canonical 1D ODE with high-frequency component and 2D convection-diffusion equation with V-cycle training strategy.
Finally, the current method is employed for the classical benchmark problem of steady Lid-driven cavity flows at different Reynolds numbers($Re$), to investigate the applicability and efficacy for the problem involved multiple modes of high and low frequency.
By virtue of various training sequence modes, improvement through predictions lead to 30$\%$ to 60$\%$ accuracy improvement.
We also investigate the synergies between current method and transfer learning techniques for more challenging problems (i.e., higher $Re$). From the present results, it also revealed that the current framework can produce good predictions even for the case of $Re = 5000$, demonstrating the ability to solve complex high-frequency PDEs.
\end{abstract}

\maketitle 


\section{Introduction}\label{sec:level1}

As machine learning continues to advance rapidly, a specific form known as Neural Networks (NN)\cite{LEE1990110,MEADE199419} has become a popular tool for solving various physics problems. When NNs are specifically designed to solve partial differential equations (PDEs) that express physical phenomena, they are referred to as Physics-Informed Neural Networks (PINNs)\cite{Raissi01,Raissi02,Raissi03}. When employ PINNs for solving PDEs, it offer several advantages, such as ability to acquire gradients via automatic differentiation\cite{AD_survey}, ability to infer the parameters from limited data and ability to train the model with less or zero data, to name a few. and through stochastic gradient descent optimizations\cite{Optimization_ML,DL}. Over the past decade, PINNs have emerged as a key research focus in diverse fields, such as structure mechanics\cite{Engineering01}, biology\cite{Biology01}, fluid mechanics\cite{CAN-PINN}, electromagnetic\cite{Maxwell01}, forecasting solutions\cite{Forecast} and assisting numerical solutions\cite{fPINN,Yuan-Ting} where they have become valuable tools for prediction and simulation.\par

In the meantime, several key areas of interest have emerged for using PINNs to solve parametric PDEs in fluid simulations. the investigated topics in this domain include solving large problems through domain decomposition\cite{cPINN,XPINN}, predicting solutions in numerical discretization forms such as large-eddy simulation (LES)\cite{DPM,ML_accel,Subgrid_LES} and finite difference methods\cite{FD}, and eliminating losses related to boundary and geometry conditions\cite{Exact_boundary, Exact_Dirichlet, LSA-PINN}, to name a few. In terms of training methods, transfer learning\cite{Transfer_fracture,Transfer_multi}, meta-learning\cite{metalearning}, and curriculum learning\cite{Failure_modes, Experts_Guide} which leverage similarities between complex problems and simpler ones, allowing models to pre-learn features and enhance accuracy by utilizing parameters from simpler questions, also been intensively studied.\par 

To improve the training,
there are two main issues to be take care with: gradient flow pathology\cite{Understanding_and_mitigating_gradients} and spectral biases\cite{Spectral}. Gradient pathology refers to issues arising from the numerical stiffness of the problem, especially the lack of sufficient collocation points needed for a stable solution. To mitigate this issue, one may employ suitable weightings in loss calculations, with adaptive weightings\cite{Understanding_and_mitigating_gradients, NTK_PINN, multitask, Self_adaptive}. Another advantage of employing the adaptive weightings is that they avoid reliance on trial and error, with the trade-off of slightly affect training speed. In this paper, we proposed an alternative way to balance accuracy and computational efficiency by performing the dimensional analysis beforehand to manually set the weightings.\par

Spectral bias refers to the tendency of neural networks to prioritize learning low-frequency functions. This issue can be mitigated by increasing the frequency of the input manifold. Tancik et al.\cite{Fourier_features, Eigenvector} introduced Fourier feature mapping, which incorporates sinusoidal functions to enhance the model's capability to predict higher frequency functions from the perspective of the Neural Tangent Kernel (NTK)\cite{NTK}, yielding significant improvements in solutions\cite{Kolmogorov}. Liu et al.\cite{MDNN} proposed a Multi-scale Deep Neural Network (MscaleDNN) structure that employs multiple sub-networks combined with supporting activation functions, enabling the model to learn the target function across different frequency scales and demonstrating improvements in solving the Poisson and Poisson-Boltzmann equations.\par

On the other hand,
It have been the intensively studies in scientific computing community that multi-grid methods can serve as a preconditioner for Fourier analysis-related frequencies\cite{FourierMG}. Although this approach has been implemented in some neural networks\cite{DNN_MG01,DNN_MG02}, its mainly usage is still limited to the numerical solvers. There are some dataset setting methods which have explored dataset establishment techniques\cite{Importance_sampling,Adaptive_sampling}.
However, their studies mainly focus on a single sample dataset with specific sampling methodologies rather than modifying dataset loading.
It is then to motivated us to address the importance the spectral bias problem 
and explore the practical usage of the multi-grid method.

Different from the study of Riccieti et al.\cite{multilevel} who previously implemented the mutil-grid method via multiple sub-networks to improve convergence rates for solving non-linear equations,
the current proposed framework is to combine concepts which inspired 
by both the multi-grid method and MscaleDNNs, as new training procedure. 
The fundamental idea of the current approach is to involve the use of varying datasets during training through transfer learning techniques, so as to mitigate the spectral bias problem. The proposed approach not only leverages dataset variation to expose the model to different frequency scales, but enhances its ability to learn target functions effectively. 
To emphasize the frequency diversity across datasets, different optimizers and schedulers are applied to each dataset. To validate our approach, we first conducted investigations by solving a high-frequency ODE, followed by a canonical convection-diffusion equation, and subsequently, we investigated different training sequence modes on the classical benchmark computational fluid dynamic problem : lid-driven cavity problem. Through these experiments, we observed improved accuracy using our proposed framework compared to conventional methods that rely on a single dataset.

This paper is structured as follows: In Section \ref{method}, we present an overview of the multi-dataset training method employed in our model, discussing how variations across datasets contribute to enhanced accuracy. This section includes flowcharts and detailed model settings. Section \ref{Results} validates our approach by comparing it to the base model with different training sequence modes through predictions of an 1D high-frequency ODE and 2D convection-diffusion equation, followed by investigating and assessing the accuracy of and efficiency of the current approach for the 2D Lid-driven cavity flow problem . 
Finally, Section \ref{conclusion} draw the conclusions, addresses current challenges, and outlines potential directions for further methodological improvements.

\section{Methodology}\label{method}

In this section, a brief overview of our multi-dataset training method with exploration of the multilevel and multi-grid methods from a frequency perspective will be described. Subsequently,  the current state of PINN models and their implementations will be proposed, together with the method of the weighting method which inspired by the dimensional analysis.

\subsection{Multi-scale NNs and Multi-grid method in frequency perspective}\label{sec2.1}
    Rahaman et al. \cite{Spectral} analyzed the Fourier spectrum within neural networks, focusing on how activation functions of neurons tend to prioritize learning simpler aspects of problems. Their experiments demonstrated a bias towards lower complexity. To counteract this, Liu et al. \cite{MDNN} introduced a multi-scale structure comprising multiple sub-networks that use inputs with varying multipliers to learn across different frequencies. On the other hand,
    Fourier frequency issues are commonly addressed by multi-grid method \cite{FourierMG} in numerical analysis society. Despite the contrasting goals—PINNs aim to establish frequencies, multi-grid methods aim to mitigate them.
    Both approaches share a common idea that the spectral bias of smaller collocation sets results in lower training frequencies, while higher frequencies are more attainable with increased collocations. Inspired by MscaleDNNs, in this study we propose an approach akin to the multi-grid method, where we dynamically switch datasets to counter both low and high frequencies directly, so as to enhance learning frequencies using different datasets.\par
    To evaluate the effectiveness of different training sequences,
    in this study we investigated our approach by ranging from a three-level coarse-to-fine mesh approach, to sequences resembling the procedure of multi-grid method, such as the popular V-cycle formulation transitioning from fine to coarse to fine meshes. 
    To efficiently implement the current methods, in this study we integrate them into a training procedure akin to transfer learning \cite{transfer_survey,transfer_torrey}, summarized as follows: (1) Establish multiple datasets, each containing varying numbers of collocation points. (2) Develop a predictive training model. (3) Sequence the training using these datasets, employing distinct optimizers and learning rate schedules individually for each dataset during training, and connecting the training progress through transfer learning principles.
    
\subsection{Physics-Informed Neural Network(PINNs)}\label{sec2.2}
    In this section,
    a brief overview of the calculation procedure in Physics-Informed Neural Networks (PINNs) will be discussed, followed by the implementations of our model, including multi-level dataset structures and settings. Finally,
    a summary of our current neural network architecture and related settings will be provided.
    
    \textbf{Overview of PINN calculation:} With the pioneer work of Karniadakis et al.\cite{Raissi01,Raissi02,Raissi03},
    physics-informed neural networks (PINNs) provide an alternative methodology to traditional numerical discretizations for solving parametric PDEs with physical interpretations. The typical PINNs employ a multi-layer perceptron (MLP) structure, take spatial inputs \textit{x} \(\in\) \(\Omega\) and temporal inputs \textit{t} \(\in\) (0,T] to predict an output \(\hat{u}\)(\textit{x, t}; \(\theta\))  that satisfies specified PDEs. This general framework can be described as follows\cite{Raissi01,CAN-PINN}:
    \begin{subequations}
    \begin{equation}\label{eq1a}
        \mathcal{N}_{t}[\hat{u}(x,t)]+\mathcal{N}_{x}[\hat{u}(x,t)]=0,  \textit{x}\in\Omega,  \textit{t} \in (0,T]
    \end{equation}
    \begin{equation}\label{eq1b}
        \mathcal{B}[\hat{u}(x,t)] = g(x,t),  \textit{x}\in\Omega,  \textit{t} \in (0,T]
    \end{equation}
    \begin{equation}\label{eq1c}
        \hat{u}(x,0) = u_o(x),  \textit{x}\in\Omega
    \end{equation}
    \end{subequations}
    where \(\mathcal{N}_{t}[\cdot]\) and \(\mathcal{N}_{x}[\cdot]\) denote differential operators for temporal and spatial derivatives, such as \textit{$u_x$}(x,t), \textit{$u_{xx}$}(x,t), \textit{$u_t$}(x,t) representing first-order spatial derivatives, second-order spatial derivatives, and temporal derivatives, respectively. \(\mathcal{B}[\cdot]\)  represents the boundary condition form, such as Dirichlet, Neumann, Robin, and with $g(x,t)$ represent the given boundary conditions.  \textit{$u_0$}(x) is represents \textit{$u$} at t = 0 which specifies initial condition.\par
    
    To generate predictive output from input \(x_{0}=(x,t)\), each neuron in Neural Network incorporate with following function:
    \begin{equation}\label{eq2}
       f_{j}(x_{0})=\sigma(w_{j}x_{0}+b_{j}),       j=1,..., k +1
    \end{equation}
    where (\(f_{j}\in \mathbb{R}^{j}\)) are hidden states, (\(\sigma\)) denotes activation functions that transform input components, and (\(w_{j}, b_{j}\)) represent weights and biases in neurons, with (\(k\)) being the total number of neurons in the layer of the Neural Network. Using the predictive output, we calculate the residuals from the PDE and errors from boundary conditions. A model loss (\(\mathcal{L}\)) is then formulated and used to update the model parameters (\(\theta\)) in current epoch (\(n\)) which comprises sets of weights 
    (\(\textbf{w}_{n}\)) and bias (\(\textbf{b}_{n}\)) for each neuron in the structure. This process can be expressed as follows:
    \begin{equation}\label{eq3}
        \theta_{n+1}=\theta_{n}-\eta \nabla_{\theta} \mathcal{L}(\theta_{n})
    \end{equation}
    
     In the formulation of PINNs, the loss function typically comprises three primary components \(\mathcal{L}_{DE}\), \(\mathcal{L}_{BC}\) and \(\mathcal{L}_{IC}\). These components correspond to the residuals of differential equations, boundary conditions, and initial conditions, respectively. The general representation of the loss function is as follows:
    \begin{equation}\label{eq4}
        \mathcal{L} \equiv \lambda_{DE}\mathcal{L}_{DE}+\lambda_{BC}\mathcal{L}_{BC}+\lambda_{IC}\mathcal{L}_{IC}
    \end{equation}
    This equation represents the weighted combination of these three fundamental components, where \(\lambda\) denotes the weighting factor assigned to each component, indicating their relative importance within the overall loss function.
    \par
    In this paper, the Mean Square Error (MSE) is used as the common loss evaluation function for each loss component:
    \begin{equation}\label{eq5}
        MSE(\hat{u},u) = \frac{1}{n}\sum^n_{i=1}(\hat{u}_i-u_i)^2
    \end{equation}
    where $\hat{u}_i$ is representing model output, and $u_i$ is corresponding target value, e.g., ground truth. With this function we can define loss components in (eq. \ref{eq4}) from (eq.\ref{eq1a}-\ref{eq1c}) to become following equations:
    \begin{subequations}
    \begin{equation}\label{eq5a}
        \mathcal{L}_{DE}=MSE( \hat{u}_{t}[ \cdot; \theta]+\mathcal{N}_{x}[\hat{u}( \cdot;\theta)], 0)
    \end{equation}
    \begin{equation}\label{eq5b}
        \mathcal{L}_{BC}=MSE(\mathcal{B}[\hat{u}(\cdot;\theta)] - g(\cdot), 0)
    \end{equation}
    \begin{equation}\label{eq5c}
        \mathcal{L}_{IC}=MSE(u(\cdot,0)-u_o, 0)
    \end{equation}
    \end{subequations}
    \par
    \par
     \textbf{Neural Network Implementations:} In this study, the Multilayer Perceptron (MLP) structure is employed. To obtain optimize training model, several crucial settings are considered in this paper:
    \begin{itemize}
        \item Activation functions: While the tanh activation function is commonly used in PINNs\cite{Self_adaptive,metalearning}, recent studies\cite{Sine_activation,Eigenvector} have shown improved accuracy with the sine activation function as a feature mapping for inputs. Therefore, the sine activation function is chosen for the first layer of the neural network. Additionally, the number of neurons in the first layer is significantly larger than in the inner layers, as larger weight distributions in the first layer help escape local minima. For the inner layers, Swish functions are employed, as previous research\cite{Activation_functions} indicates that this function is less susceptible to vanishing and exploding gradient problems.\par  
        \item Weight and Bias Initialization: The initialization of weights and biases is crucial for network training. In this paper, the weights of the first layer , particularly the sine activation functions, are initialized from a normal distribution \(\mathcal{N}\)(0, \(\sigma^2\)), \(\sigma\) =1. The weights of subsequent layers are initialized using the He uniform initialization\cite{Kaiming, NEURIPS2020_53c04118} or the similar Xavier uniform initialization, as the setting has been proofed to be more effective \cite{Sine_activation}, while the biases are initialized from a uniform distribution, following the default settings in PyTorch\cite{PyTorch}.
        \item Optimizer and Scheduler Settings: The commonly used optimizer is Adam\cite{adam}, a variant of stochastic gradient descent. It is accompanied by the ReduceLROnPlateau or StepLR scheduler, which dynamically adjusts the learning rate based on the monitored loss value. This adaptive learning rate adjustment enhances model convergence rate and improves training efficiency.
    \end{itemize}
    
\subsection{Dimensional analysis weighting method root scheme($DW_{root}$)}\label{sec2.3}
    The Dimensional analysis weighting method was introduced by Chou et al. \cite{YiEn}. This method involves analyzing the order of magnitude for each component within the loss functions. The primary concept behind this method is to establish a balance among the weights of different components by considering their ratios, ensuring each component carries a similar level of significance. The method implementation is presented in eq.\ref{eq10} and eq.\ref{eq11}. This method serves as an alternative way to adapt weighting and offer two main advantages: (1) Good convergence throughout the training sequence and (2) No need for additional computations within the model. For the sake of completeness,
the proposed algorithm of this paper is summarized in Alg.\ref{alg. 1}
    \begin{algorithm}\caption{Multi-dataset PINN}\label{alg. 1}
        \textbf{Step 1:}
            Set up input matrices and model parameters \;
        \textbf{Step 2:}
            Construct a Fully connected PINN model with initialized weights and bias (for \(k_{i}\)-th iteration represented as \(\textit{w}_k\))\;
        \textbf{Step 3:}
            Declare dimensional analyzed weights\;
        \textbf{Step 4:}
            Setup optimizers and schedulers for different dataset level\;
        \textbf{Step 5:}
            Do \textit{K} gradient descent iterations in each level defined level sequence to update neural network:\;
        \For{\textit{$Level = 1$ to $N-1$}}{
        \For{\textit{$k=1$ to $K$}}{
            {Tune neural network weights and bias with respect to \(\mathcal{L}\)\;
            \(w_{k+1}\gets\mathcal{L}(w_k)\)
            }
        }}
        \textbf{Step 6:} Train the last training level
        \For{\textit{$k=(N-1)*K$ to $K_{max}$}}{
            Tune neural network weights and bias with respect to \(\mathcal{L}\)\;
            \(w_{k+1}\gets\mathcal{L}(w_k)\)
        }
    \end{algorithm}

\section{Results and Discussion}\label{Results}
In this section, the proposed method is demonstrated through various training sequence modes that involve multiple dataset levels to showcase improvements. The examples below illustrate test problems that highlight the efficacy of the method: (i) solve a high-frequency sine function to demonstrate its capability in addressing the spectral bias problem; (ii) investigate different training sequence modes via 2D Smith-Hutton problem to reveal the accuracy and efficiency of current methods; (iii) investigate the 2D steady Lid-driven cavity flow to further demonstrate the applicability, accuracy and efficiency for the challenging system of partial differential equations, e.g., Navier-Stokes equations. The limitations with specified networks and datasets, followed by comparisons of the proposed model's accuracy with results from various studies will also be examined.

\subsection{Experimental setup}\label{sec3.1}
For 1D ODE problem , A 2-level random collocation set using 16 and 64 points will be studied. For 2D problems, 3-level uniform collocation sets employ 435, 1770, and 6670 points for Smith-Hutton problem, while different random collocation point sets for lid-driven cavity flow problem are employed, respectively, as shown in Table. \ref{table3.1} and \ref{table3.1-2}. Each collocation set in this paper separates inner points and boundary points for training: inner points are used for calculating the residuals of differential equations (\(\mathcal{L}_{DE}\) ), while boundary points satisfy the boundary conditions  (\(\mathcal{L}_{BC}\)).  Details of which will be provided in subsequent sections. All PINN structures utilize a fully connected feed-forward architecture with activation functions shown separately in sections. The width and depth of the neural networks are tailored for each experiment to achieve reasonable results while minimizing training costs from excessive model parameters. Model accuracy is assessed using the Relative \(L^{2}\) error (Rel. $L_2$) metric, defined as follows:
\begin{equation}\label{eq6}
        \frac{||u-\hat{u}||_{2}}{||u||_{2}}
    \end{equation}
For all experiments, we used the Adam optimizer with full batch training,
along with either the \textit{StepLR} scheduler or the \textit{ReduceLROnPlateau} scheduler, both provided by PyTorch. All experiments were conducted on a system equipped with an Intel i9-13900K processor and a NVIDIA RTX 4070Ti GPU with 12GB of VRAM.
    
 \begin{table}[htbp!]
        \centering
        \resizebox{1\textwidth}{!}{%
        \begin{tabular}{p{5cm} | p{6cm} p{6cm} }
        Problem & High frequency \newline ODE (sec.\ref{sec3.2}) & Smith-Hutton problem (sec.\ref{sec.SH}) \newline Pe = 1000\\
        Govern eq. &  eq.\ref{eq7b} & eq.\ref{eq-sh} \\
        \hline
        Neural Network structure & (x)-64-64-64-64-($\hat{u}$) & (x,y)-128-64-64-64-($\hat{u}$) \\
        Training collocation points & 64+32 & 6670+1770+435 \\
        Initial learning rate & 1e-3 & 1e-3 \\ 
        Max Training epochs & 30,000 & 50,000 \\
        \hline
        \end{tabular}%
        }
        \newline
        \\Above table shows model setting by default, which varies for different tests \newline with details shown in Appendix (sec.\ref{App}).
        \caption{Experiments and default setups for 1D ODE and 2D Smith-Hutton problems}
        \raggedright
        \label{table3.1}
    \end{table}

 \begin{table}[htbp!]
        \centering
        \resizebox{1\textwidth}{!}{%
        \begin{tabular}{p{5.5cm} | p{4cm} p{4cm} p{4cm} }
        Problem & Lid-driven cavity \newline(sec.\ref{sec.3.3}) \newline Re=400 & Lid-driven cavity \newline(sec.\ref{sec.3.3})\newline Re=1000& Lid-driven cavity \newline(sec.\ref{sec.3.3})\newline Re=5000 \\
        Govern eq. &  eq.\ref{eq9} & eq.\ref{eq9} & eq.\ref{eq9} \\
        \hline
        Neural Network structure & (x,y)-64-20-20-[(20-20-($\hat{\psi}$))-(20-20-($\hat{p}$))] & (x,y)-256-64-64-[(64-64-($\hat{\psi}$))-(64-64-($\hat{p}$))]&  (x,y)-256-64-64-[(64-64-($\hat{\psi}$))-(64-64-($\hat{p}$))]\\
        Training collocation points & 6565+1685+445 & 6565+1685+445 &  14645
+3725+965 \\
        Initial learning rate & 1e-3 & 1e-3 & 1e-3  \\ 
        Max Training epochs & 200,000\newline/300,000(w-cycle)\newline/200,000(V-cycle) & 212,000(V-cycle) & 1,207,000 (curriculum)\\
        \hline
        \end{tabular}%
        }
        \newline
        \\Above table shows model setting by default, which varies for different tests \newline with details shown in Appendix (sec.\ref{App}).
        \caption{Experiments and default setups for Lid-driven cavity problem}
        \raggedright
        \label{table3.1-2}
    \end{table}

\subsection{1D high frequency ODE}\label{sec3.2}
In order to demonstrate the effectiveness of our method in addressing the aforementioned spectral bias issue, we consider the following 1D sine function:
\begin{equation}\label{eq7a}
    f(x)=sin(10x),x\in[0,2\pi]
\end{equation}
which represents a high frequency ordinary differential equation (ODE), expressed as:
\begin{equation}\label{eq7b}
    g(x)=\frac{\partial f(x)}{\partial x}=10sin(10x)
\end{equation}
where \(g(x)\) represents the residual equation for the differential equation loss (\(\mathcal{L}_{DE}\)), and \(f(0)=0, f(2\pi)=0\)  denote boundary conditions for the boundary loss (\(\mathcal{L}_{BC}\)).\par
In this experiment, we set the weighting to unity for clarity of observation. The dataset used in this section, denoted as (\(D_1\)), which is main collocation in this section , consists of 64 randomly sampled collocation points and is primarily used for training to satisfy the equation. Additionally, a testing set containing 256 randomly sampled points is employed to evaluate the model's accuracy against ground truth over 30,000 Adam epoch. The model structure consists of a 4-layer MLP with a width of 64. Activation functions are expressed as follows:
\begin{equation}\label{eq8}
    \sigma_{l}(x)= 
    \begin{cases}
      sin(2\pi x), & l=1\\
      sin(x), & 2\le l \le 4\\
    \end{cases}
\end{equation}
where \(l\) is the layer number. The He uniform initialization is employed, while the biases for this problem is set as zero.
\par
To compare training with only main dataset(\(D_1\)) for 30,000 epochs, we introduce an additional auxiliary dataset (\(D_2\) in this section) with 32 random samples, to assist \(D_1\) in converge. Training of the additional dataset will interrupt the training of \(D_1\), which result in following training sequence (Table. \ref{table3.1}).\par
    \begin{table}[htbp!]
        \centering
        \resizebox{.6\textwidth}{!}{%
        \begin{tabular}{p{3cm} p{3cm} p{4cm}}
        Level & Dataset & training epochs  \\
        \hline
        1 & \(D_1\) &  1$\le$ n $\le$ 5,000\\
        2 & \(D_2\) & 5,001$\le$ n $\le$ 15,000 \\ 
        3 & \(D_1\) &  15,001$\le$ n $\le$ 30,000\\
        \hline
        \end{tabular}%
        }
        \caption{ODE training sequence}
        \raggedright
        \label{table3.2}
    \end{table}
    During the training of \(D_2\), we employ different optimizers and schedulers to facilitate the model's learning of additional frequencies within a shorter training period. This training sequence is structured with the following objectives: the first level establishes the model's recognition of the target dataset, the second level focuses on learning new frequencies, and the final level integrates both learned frequencies. The detailed setting is shown in Appendix.(sec\ref{App.1}).
    
    \begin{figure}[htbp!]
    \captionsetup{justification=centering}
        \begin{subfigure}{0.48\textwidth}
            \includegraphics[width=1\linewidth, height=5cm]
            {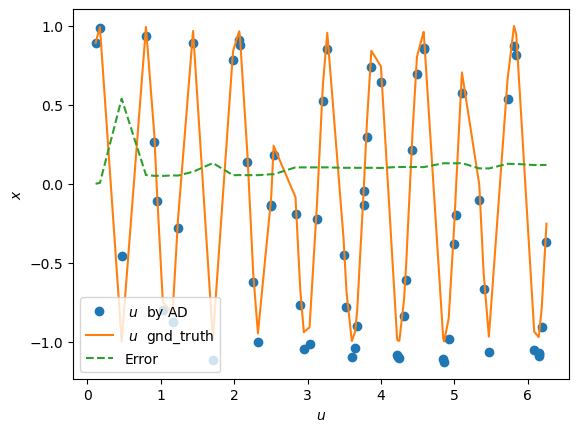} 
            \caption{PINN w/o proposed method}
            \label{fig3.1a}
            \end{subfigure}
        \begin{subfigure}{0.48\textwidth}
            \includegraphics[width=1\linewidth, height=5cm]
            {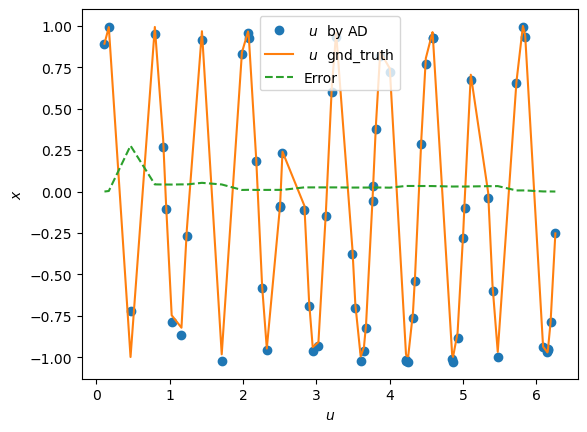}
            \caption{PINN w/ proposed method}
            \label{fig3.1b}
        \end{subfigure}
        \caption{Training result (blue dot), Training ground truth (solid line) and Absolute error (dashed line) after end training for 30,000 epochs}
    \label{fig3.1}
    \end{figure}
    
    \begin{figure}[htbp!]
    \captionsetup{justification=centering}
        \begin{subfigure}{0.48\textwidth}
            \includegraphics[width=1\linewidth, height=5cm]
            {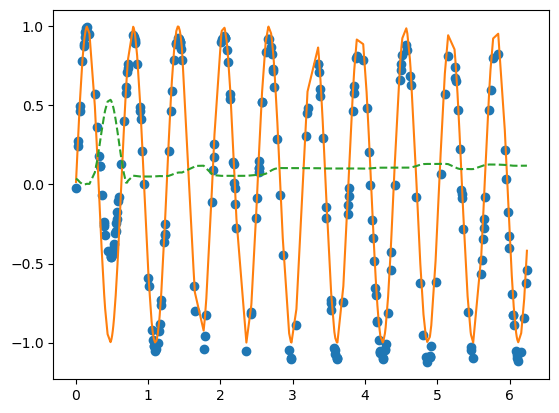} 
            \caption{PINN w/o proposed method}
            \label{fig3.2a}
            \end{subfigure}
        \begin{subfigure}{0.48\textwidth}
            \includegraphics[width=1\linewidth, height=5cm]
            {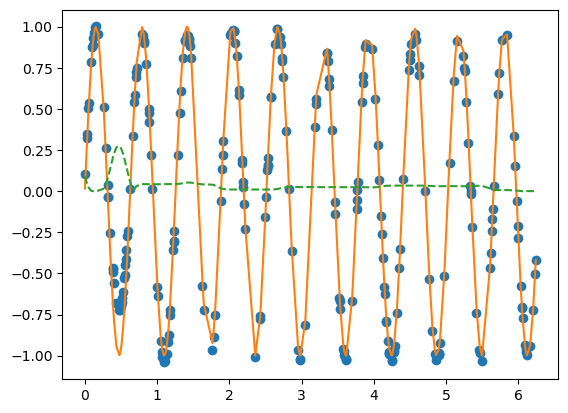}
            \caption{PINN w/ proposed method}
            \label{fig3.2b}
        \end{subfigure}
    \caption{Test result (blue dot), Test ground truth (solid line) and Absolute error (dashed line) after end training for 30,000 epochs}
    \label{fig3.2}
    \end{figure}
    
    \begin{table}[htbp!]
        \centering
        \resizebox{1\textwidth}{!}{%
        \begin{tabular}{p{5cm} | p{3cm} p{3cm} p{3cm} p{3cm}}
        Model & \(\mathcal{L}_{DE}\) & \(\mathcal{L}_{BC}\) & Time cost(sec) & Test error(Rel. $L_2$)  \\
        \hline
        PINN w/o proposed method & 5.518e-07 & 2.906e-12 & 164 & 1.15e-01\\
        PINN w/ proposed method & 1.863e-06 & 1.120e-10 & 169 & 4.16e-02\\ 
        \hline
        \end{tabular}%
        }
        \caption{ODE(sec.\ref{sec3.2}) Loss values comparison after 30,000 epochs of training}
        \raggedright
        \label{table3.3}
    \end{table}
As shown in Table. \ref{table3.3} and Figures \ref{fig3.1} and \ref{fig3.2}, PINNs often struggle to solve high-frequency ODEs despite achieving low loss values. As our method resolve this issue by incorporating additional smaller datasets within the training sequence, which incurs a modest increase in training time but improves performance.

\subsection{2D Smith-Hutton problem}\label{sec.SH}
To further demonstrate the applicability and efficiency , another benchmark problem, the Smith-Hutton problem,
is investigated. In this study, the scalar field is governed by the 2D convection diffusion equation,
and driven by the following velocity field in the domain of $(x,y)=(-1\sim1,0\sim1)$\cite{Smith82}:
\begin{subequations}
\begin{align}
&\frac{\partial \phi}{\partial x} + v\frac{\partial \phi}{\partial y} - \frac{1}{Pe}  (\frac{\partial^2 \phi}{\partial x^2}+\frac{\partial^2 \phi}{\partial x^y}) = 0\\
u &=2y(1-x^2)\\
v &=-2x(1-y^2)
\end{align}
\end{subequations}
where the $Pe$ is the Pelect number and chosen as 1000 in this study.
The top, left and right boundary are set as $\phi=1-tanh(10)$,
and the bottom boundary is set as:
\begin{subequations}
\label{eq-sh}
\begin{align}
\phi =1+tanh(10(2x+1)), &\quad x\le 0\\
\frac{\partial \phi}{\partial y} = 0, &\quad x >0
\end{align}
\end{subequations}
The loss function for the current investigated problem is defined as follows:
\begin{align}
\mathcal{L}_{DE}\ &= u\frac{\partial \phi}{\partial x} + v\frac{\partial \phi}{\partial y} - \frac{1}{Pe} (\frac{\partial^2 \phi}{\partial x^2}+\frac{\partial^2 \phi}{\partial x^y})\\
\mathcal{L}_{DBC}\ &= \phi - (1+tanh(10(2x+1)))\\
\mathcal{L}_{NBC}\ &= \frac{\partial \phi}{\partial y}
\end{align}
in the above, $\mathcal{L}_{DE}$, $\mathcal{L}_{DBC}$ and $\mathcal{L}_{NBC}$ are the loss components for the governing equation, Dirichlet boundary condition and Neumann boundary condition. The weight for each component is chosen as 1 in this investigated problem.

To demonstrate the efficiency of the current method,
the training is performed with conventional training and V cycle training procedure.
for the conventional training,
the uniform sample points of 6670 is employed for the training with 50,000 epochs,
while for the V cycle method,
the three level of sample points, 6670 ($D_1$),1770 ($D_2$)and 435 ($D_3$) is employed and trained with 10,000 epochs at each level,
which lead to 50,000 epochs in total.
For the first hidden layer,
sine activation function is employed,
while for other hidden-layers we choose tanh as activation function.
The details of the training procedure for V cycle is tabulated in table.\ref{table-sh}.
    \begin{table}[htbp!]
        \centering
        \resizebox{.6\textwidth}{!}{%
        \begin{tabular}{p{3cm} p{3cm} p{4cm}}
        Level & Dataset & training epochs  \\
        \hline
        1 & \(D_1\) &  1$\le$ n $\le$ 10,000\\
        2 & \(D_2\) & 10,001$\le$ n $\le$ 20,000 \\ 
        3 & \(D_3\) &  20,001$\le$ n $\le$ 30,000\\
        4 & \(D_2\) &  30,001$\le$ n $\le$ 40,000\\
        5 & \(D_1\) &  40,001$\le$ n $\le$ 50,000\\
        \hline
        \end{tabular}%
        }
        \caption{V cycle training sequence for Smith-Hutton problem}
        \raggedright
        \label{table-sh}
    \end{table}

\begin{figure}[htbp!]
    \centering
    \captionsetup{justification=centering}
    \includegraphics[width=1\linewidth, height=7cm]{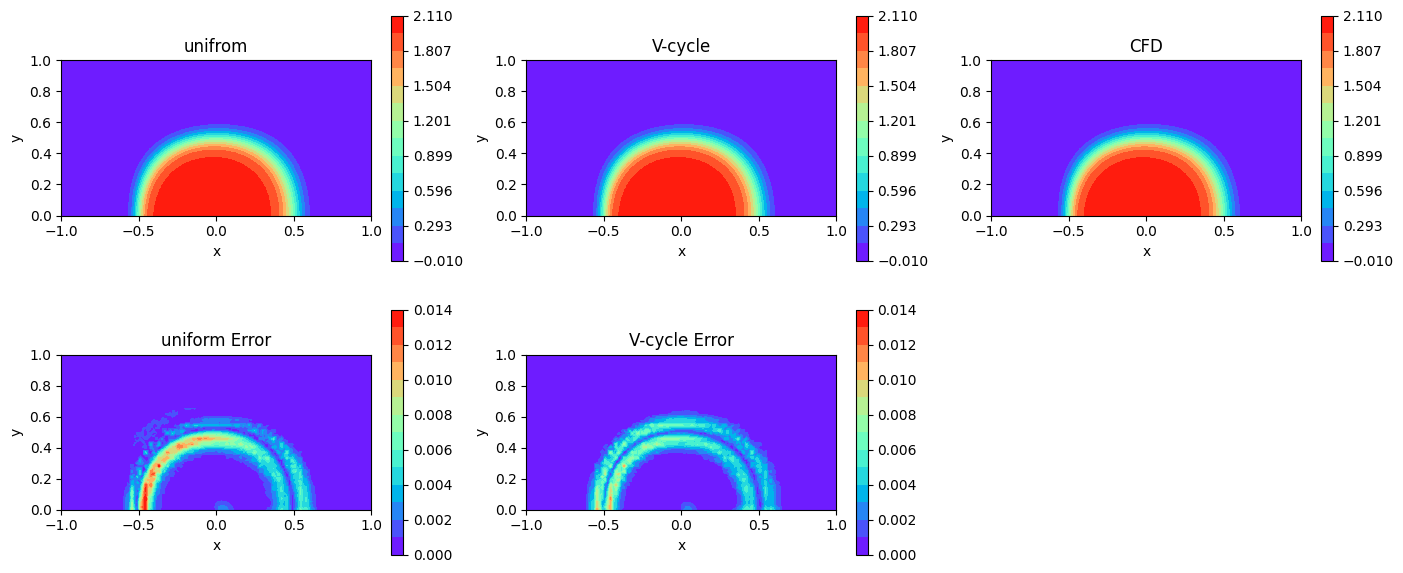} 
    \caption{Comparisons of training result (top) and absolute error (bottom) for different training settings.}
    \label{fig:sh}
    \end{figure}

From Fig. \ref{fig:sh} and Table \ref{table-sh-error},
it can be seen that the current proposed method improve the accuracy of the training results.
By comparing the the relative L2-norm with the CFD reference solutions as ground truth,
the proposed method can improve the accuracy of 20\%. The applicability and accuracy of the current method is confirmed for 2D Smith-Hutton problem.

    \begin{table}[htbp!]
        \centering
        \resizebox{1\textwidth}{!}{%
        \begin{tabular}{p{5cm} | p{3cm} p{3cm} p{3cm} p{3cm} }
        Model & \(\mathcal{L}_{DE}\) & \(\mathcal{L}_{DBC}\) & \(\mathcal{L}_{NBC}\) & Test error(Rel. $L_2$)  \\
        \hline
        uniform & 3.54e-06 & 6.33e-07 & 1.49e-08 & 2.54e-03\\
        V cycle & 5.10e-06 & 7.87e-09 & 3.70e-08 & 2.10e-03\\ 
        \hline
        \end{tabular}%
        }
        \caption{Loss values comparison after 50,000 epochs of training for the Smith-Hutton problem}
        \raggedright
        \label{table-sh-error}
    \end{table}
    
\subsection{2D steady Lid-driven cavity}\label{sec.3.3}
Finally, we investigate the Lid-driven cavity problem, a classic benchmark used to assess the accuracy of methodologies in Computational Fluid Dynamics (CFD). In this study, the effect of grid sequence on model accuracy, evaluating Relative $L_2$ error(Rel. $L_2$) and Mean Absolute Error (MAE) against the reference solution by Ghia et al.\cite{GHIA1982387} will be studied.\par
\textbf{Problem description:} Under the isothermal and incompressible assumptions, 
the flow field in the Lid-driven cavity is governed by the incompressible Navier-Stokes equations, i.e., momentum equations (denotes $NS_x$ and $NS_y$) and Continuity equation (denotes $Con$):
\begin{equation}\label{eq9}
    \begin{cases}
      f_{NS_x}=u\frac{\partial u}{\partial x}+v\frac{\partial u}{\partial y}+\frac{1}{\rho}[\frac{\partial p}{\partial x}-\mu(\frac{\partial^2 u}{\partial x^2}+\frac{\partial^2 u}{\partial y^2})] \\
      f_{NS_y}=u\frac{\partial v}{\partial x}+v\frac{\partial v}{\partial y}+\frac{1}{\rho}[\frac{\partial p}{\partial y}-\mu(\frac{\partial^2 v}{\partial x^2}+\frac{\partial^2 v}{\partial y^2})]\\
      f_{Con}=\frac{\partial u}{\partial x}+\frac{\partial v}{\partial y}
    \end{cases}
\end{equation}
With top boundary $U_{lid} = 1$, density $\rho=1$ and domain $\Omega(x,y) \in [0, 1]\times[0,1]$, 
the Reynolds number (Re) is then be defined as $Re=\frac{\rho U_{lid}}{\mu}$.\par
This is problem is well-known to be difficult when higher $Re$ is used.
To model this challenging problem,
Wang et al. \cite{Understanding_and_mitigating_gradients} addressed gradient pathology issues and employing adaptive loss weights. Li et al. \cite{li2022dynamic} proposed an adaptive loss weighting approach optimized through neural networks. Karniadakis et al. \cite{karniadakis2023solution} introduced a two-network structure incorporating parameters for eddy viscosity and entropy equations. Jiang et al. \cite{jiang2023applications} utilized finite difference methods for accurate gradient calculations and rapid convergence. Wang et al. \cite{Experts_Guide} provided a comprehensive review, highlighting advancements such as Fourier features \cite{Fourier_features}, Neural-Tangent-Kernel based Weighting \cite{NTK_PINN}, and curriculum training \cite{Failure_modes}. Chen et al. \cite{chen2021improved} proposed a citation block-based structure using Convolutional Neural Networks and sample weighting mechanisms. Lastly, Chiu et al. \cite{CAN-PINN} proposed a gradient calculation method combining automatic and numerical differentiation. In this study, we will show the by using a simple multilevel idea can make the training feasible.\par

To better provide the training weight for each individual loss component,
in this study we employ the dimensional-analyzed weighting scheme.

Beginning with dimensional analysis of model losses, each loss component can be expressed with dimensions as follows:
\begin{equation}\label{eq10}
    \begin{cases}
      \mathcal{L}_{NS_x}\sim(\mu(\frac{\partial^2 u}{\partial x^2}+\frac{\partial^2 u}{\partial y^2}))^2\sim\frac{[\mu]^2[U]^2}{[h]^4}\\
      \mathcal{L}_{NS_y}\sim(\mu(\frac{\partial^2 v}{\partial x^2}+\frac{\partial^2 v}{\partial y^2}))^2\sim\frac{[\mu]^2[U]^2}{[h]^4}\\
      \mathcal{L}_{Con}\sim(\frac{\partial u}{\partial x}+\frac{\partial v}{\partial y})^2\sim \frac{[U]^2}{[h]^2}\\
      \mathcal{L}_{BC_u}\sim(u-g_u)^2\sim[U]^2\\
      \mathcal{L}_{BC_v}\sim(v-g_v)^2\sim[U]^2\\
    \end{cases}
\end{equation}
where $BC$ represents boundary condition in u- and v- components, and $h$ is the specific length to be defined.\par 
To balance the losses, the corresponding weightings will have reciprocal dimensions of eq.(\ref{eq10}) after the equation eliminates repeating elements:
\begin{equation}\label{eq11}
    \begin{cases}
      \lambda_{NS_x}=\frac{[h]^4}{[\mu]^2}\\
      \lambda_{NS_y}=\frac{[h]^4}{[\mu]^2}\\
      \lambda_{Con}=[h]^2\\
      \lambda_{BC_u}= 1\\
      \lambda_{BC_v}= 1\\
    \end{cases}
    Root scheme\rightarrow
    \begin{cases}
      \lambda_{NS_x}=\frac{[h]^2}{[\mu]}\\
      \lambda_{NS_y}=\frac{[h]^2}{[\mu]}\\
      \lambda_{Con}=[h]\\
      \lambda_{BC_u}= 1\\
      \lambda_{BC_v}= 1\\
    \end{cases}
\end{equation}
In this study, $h$ is chosen as $\frac{1}{N}$ when a uniform $N\times N$ samples are employed,
while in the random collocation scenario, $h$ is determined by the closest number to square root of the number of main collocations. For example, with 6565 main collocations, we can get $h=\frac{1}{Round(\sqrt{6565})}=\frac{1}{81}$.\par

\begin{figure}[htbp!]
    \captionsetup{justification=centering}
            \includegraphics[width=.6\linewidth, height=6cm]{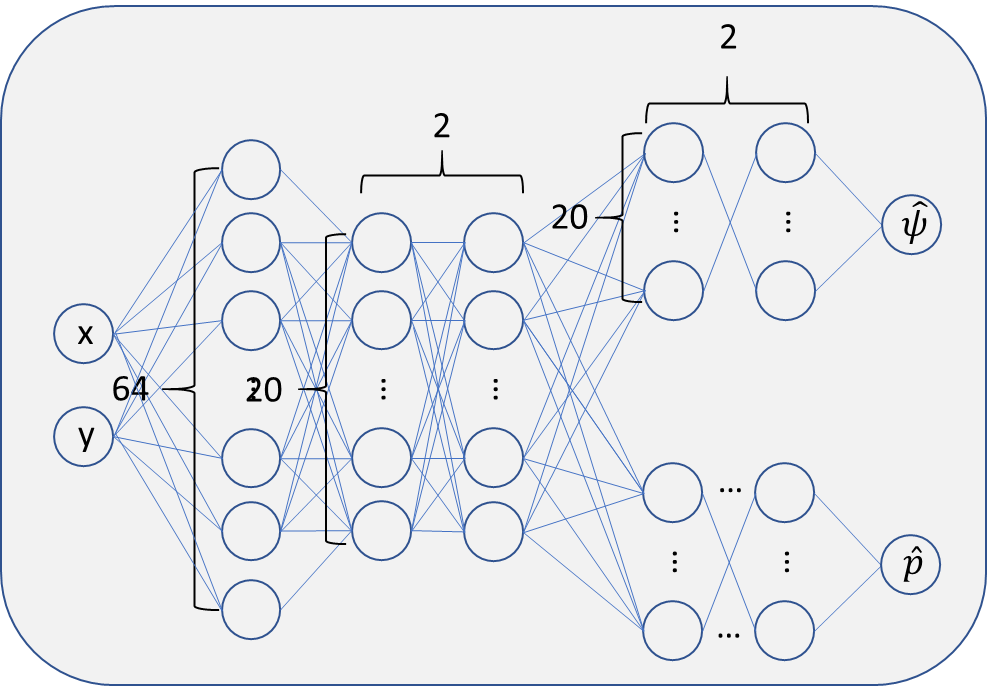} 
    \caption{NN structure for Lid-driven cavity}
    \label{fig3.3}
    \end{figure}

Our neural network is structured into two sub-networks for predicting two distinct values, which this structure is inspired by Wong et al.\cite{Sine_activation,CAN-PINN} with a decent accuracy in original structure in Lid-driven cavity, as illustrated in the figure(Fig.\ref{fig3.3}) and tabulated in Table. \ref{table3.1}, where $\hat{\psi}$ and $\hat{p}$ each utilize a sub-network with 2 layers and 20 units.
For the first hidden layer,
sine activation function is employed,
while for other hidden-layers we choose Swish as activation function:
\begin{equation}\label{eq12}
    \sigma_{l}(x)= 
    \begin{cases}
      sin(2\pi x), & l=1\\
      Swish(x)=x*sigmoid(x), & l > 1\\
    \end{cases}
\end{equation}
To ease the training,
the vorticity-velocity (VV) formulation \cite{NSFnet} is also chosen:
\begin{equation}\label{eq13}
      \hat{u}=\frac{\partial \hat{\psi}}{\partial y},  \hat{v}=-\frac{\partial \hat{\psi}}{\partial x}
\end{equation}

In every training conducted in this subsection, two convergence criteria have been employed: the Total Loss Criterion and the Loss Plateau Criterion. For the Total Loss Criterion, the following equation is employed:
    \begin{equation}\label{eq14a}
          f^{crit}_{i} = |log_{10}(\mathcal{L}^{n}_i)-log_{10}(\mathcal{L}^{n-1000}_i)|
    \end{equation}
    where $i$ is the loss components and $n$ is current epoch.
For the Loss Plateau Criterion, we use the following equation (Eq. \ref{eq14b}) to halt training if the losses remain unchanged for a specified duration
    \begin{equation}\label{eq14b}
          Crit_{Plat} := \frac{\sum_i^{N_{comp}}{f^{crit}_i}}{N_{comp}}
    \end{equation}
By default, the model stops training when either the total loss drops below  $10^{-6}$ or $Crit_{Plat}\le10^{-4}$. If neither condition is met, training stops at the maximal epochs defined in Table.\ref{table3.1}.\par  
Similar to previous experiments, several datasets are used in training. We aim to investigate the effectiveness of different dataset sequences and structures, as outlined in the Table.\ref{table3.4}.\par
\begin{table}[htbp!]
        \centering
        \resizebox{1\textwidth}{!}{%
        \begin{tabular}{p{5cm} p{5cm} | p{5cm} p{5cm}}
        Dataset & Collocations & Dataset & Collocations \\
        \hline
        \(D_1, D_{1a}\)(main) &  6565 & \(D_{avg1}\) & 2900 \\
        \(D_2, D_{2a}\) & 1685 &  \(D_{avg2}\) & 2900 \\ 
        \(D_3\) & 445 &  \(D_{avg3}\) & 2900\\
        \hline
        \end{tabular}%
        }
        \caption{Datasets detail in Lid-driven cavity}
        \raggedright
        \label{table3.4}
    \end{table}
Where $D_1-D_3$(Random) are datasets with randomly distributed collocation points. $D_{1a},D_{2a}$(Inherit) are datasets where the collocation points inherit coordinates from a preceding smaller dataset level. $D_{avg1}-D_{avg3}$(Average) are three datasets with the same number of collocations but different coordinates. Then, we perform tests with the following sequence and composition as shown in Table. \ref{table3.5}:\par
\begin{table}[htbp!]
        \centering
        \resizebox{1\textwidth}{!}{%
        \begin{tabular}{p{3cm} | p{2cm} p{2cm} p{2cm} p{2cm} p{2cm}p{2cm} p{2cm}p{2cm} p{2cm}}
        Level & 1 & 2 & 3 & 4 & 5 & 6 & 7 & 8 & 9\\
        \hline
        Duration(W-cycle) &  20,000 &  20,000 &  20,000 &  20,000 & 20,000 &  20,000 &  20,000 &  20,000 &  To converge\\
        Duration(V-cycle) &  20,000 &  20,000 &  20,000 &  20,000 &  To converge\\
        Duration(3-level) &  40,000 &  40,000 &  To converge \\
        \hline
        Random, 3 levels & $D_3$ & $D_2$ & $D_1$\\
        Inherit, 3 levels & $D_3$ & $D_{2a}$ & $D_{1a}$ \\ 
        Random, V-cycle & $D_1$ & $D_2$ & $D_3$ & $D_2$ & $D_1$\\
        Inherit, V-cycle & $D_{1a}$ & $D_{2a}$ & $D_3$ & $D_{2a}$ & $D_{1a}$\\
        Random, W-cycle & $D_{1}$ & $D_{2}$ & $D_3$ & $D_{2}$ & $D_{1}$ & $D_{2}$ & $D_3$ & $D_{2}$ & $D_{1}$ \\
        Average, V-cycle & $D_{avg1}$ & $D_{avg2}$ & $D_{avg3}$ & $D_{avg2}$ & $D_{avg1}$\\
        \hline
        \end{tabular}%
        }
        \caption{Test detail in Lid-driven cavity}
        \raggedright
        \label{table3.5}
\end{table}
 Table. \ref{table3.6} presents our training results for Re=400 ($\mu=\frac{1}{400}$) using the default setting alongside a reference solution. In addition to the aforementioned datasets, we also compare the accuracy achieved with a uniform (81$\times$81) mesh dataset which is obtained by finite difference method(FDM) with central difference scheme (CDS), and a single random dataset ($D_1$) to demonstrate the efficacy of our method. Across all tests, significant improvements in accuracy are observed, with the random V-cycle showing the largest improvement by reducing the error by 64\%.
\par

\begin{table}[htbp!]
        \centering
        \resizebox{.7\textwidth}{!}{%
        \begin{tabular}{p{3cm} | p{2cm} p{2cm} p{2cm} p{2cm}}
       Test & $Error_u$\newline(MAE) & $Error_v$\newline (MAE) & Training epochs & Error\newline decrease \\
        \hline
        Uniform & 4.48e-02 & 5.05e-02 & 100,000  \\
        Random, 1 level & 1.75e-02 & 2.90e-02 & 104,000 & baseline \\
        \hline
        Random, 3 levels & 1.10e-02 & 2.40e-02 & 180,000 & -27.9\%\\
        Inherit, 3 levels & 8.3e-03 & 1.74e-02 & 182,000 & -46.3\%\\ 
        \textbf{Random, V-cycle} & \textbf{4.25E-03} & \textbf{1.38E-02} & \textbf{163,000} & \textbf{-64.1\%}\\
        Inherit, V-cycle & 8.69e-03 & 1.97E-02 & 163,000 & -41.2\%\\
        Random, W-cycle & 1.04E-02 & 2.12E-02 & 162,000 & -33.7\%\\
        Average, V-cycle & 7.91E-03 & 1.75E-02 & 184,000 & -47.2\%\\
        \hline
        \end{tabular}%
        }
        \newline
        \newline
        Error decrease = $\frac{1}{2}\times(\frac{Error_{u,model}-Error_{u,base}}{Error_{u,base}}+\frac{Error_{v,model}-Error_{v,base}}{Error_{v,base}})$
        \caption{Best Test Result MAE compare with reference solution, $Re$ = 400}
        \raggedright
        \label{table3.6}
\end{table}
\begin{figure}[htbp!]
    \captionsetup{justification=centering}
        \begin{subfigure}{0.48\textwidth}
            \includegraphics[width=1\linewidth, height=6cm]
            {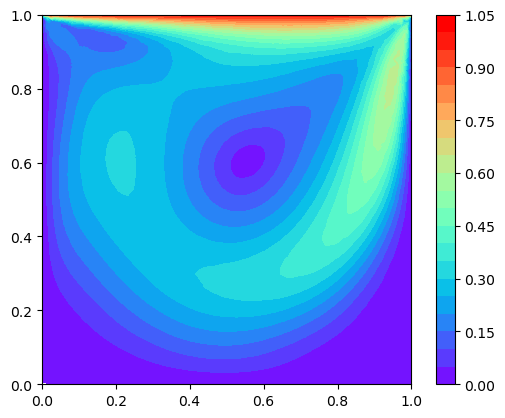} 
            \caption{}
            \label{fig3.4a}
            \end{subfigure}
        \begin{subfigure}{0.48\textwidth}
            \includegraphics[width=1\linewidth, height=6cm]
            {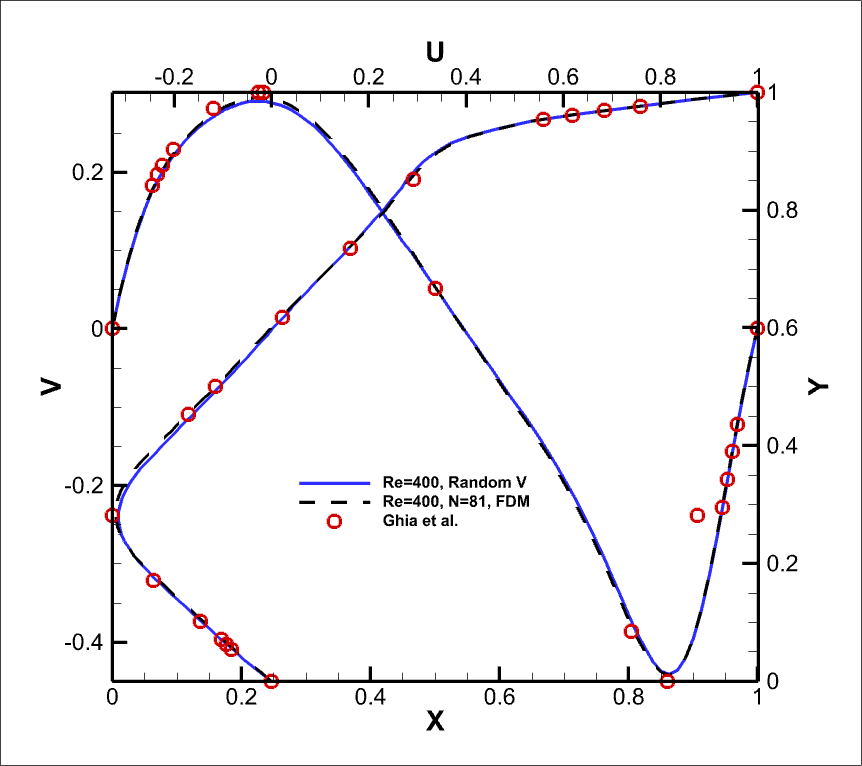}
            \caption{}
            \label{fig3.4b}
        \end{subfigure}
    \caption{(a)Re=400 Model training result\newline(b)Re=400 test result(Blue solid line), FDM($81\times81$ CDS)(Black dashed line),\newline and Ghia et al solution(Red circle) after training for 163,000 epochs with Random, V-cycle}
    \label{fig3.4}
    \end{figure}
  As the Reynolds number increases, the model becomes infeasible to converge. 
  To address this, we scaled up our PINN model by making it four times wider with hidden layers. Specifically, the first layer now contains 256 neurons, while subsequent layers have 64 neurons each. This width was chosen considering computational time constraints and for benchmarking against other studies\cite{Experts_Guide}. Without curriculum training, our model predicts Re=1000 with an accuracy of 5.85e-02 using a random V-cycle structure, as illustrated in Figure \ref{fig3.5}.\par
\begin{figure}[htbp!]
    \captionsetup{justification=centering}
        \begin{subfigure}{0.48\textwidth}
            \includegraphics[width=1\linewidth, height=6cm]{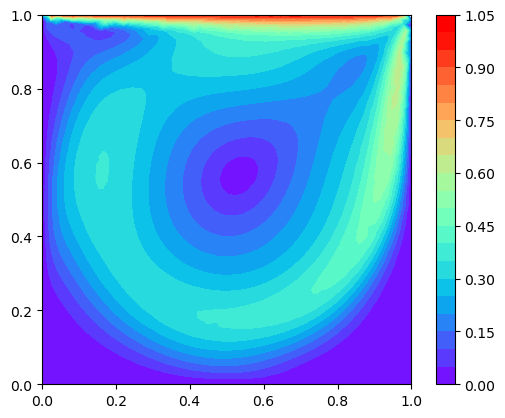} 
            \caption{}
            \label{fig3.5a}
            \end{subfigure}
        \begin{subfigure}{0.48\textwidth}
            \includegraphics[width=1\linewidth, height=6cm]{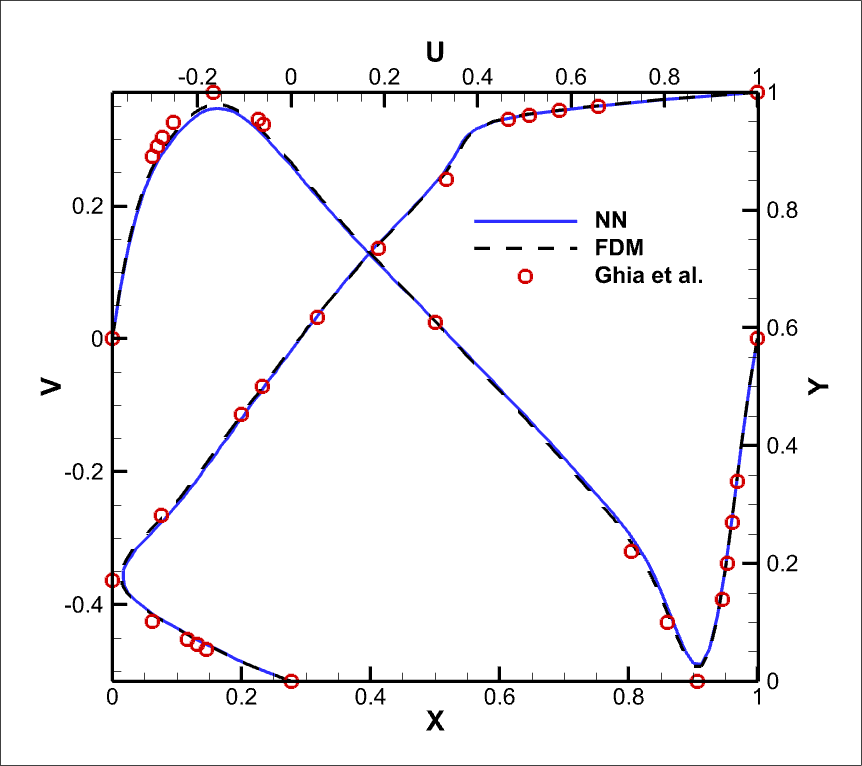}
            \caption{}
            \label{fig3.5b}
        \end{subfigure}
    \caption{(a)Re=1000 Model training result\newline(b)Re=1000 test result(Blue solid line), FDM($81\times81$ CDS)(Black dashed line),\newline and Ghia et al solution(Red circle) after training for 212,000 epochs}
    \label{fig3.5}
    \end{figure}
\begin{figure}[htbp!]
    \captionsetup{justification=centering}
        \begin{subfigure}{0.48\textwidth}
            \includegraphics[width=1\linewidth, height=6cm]
            {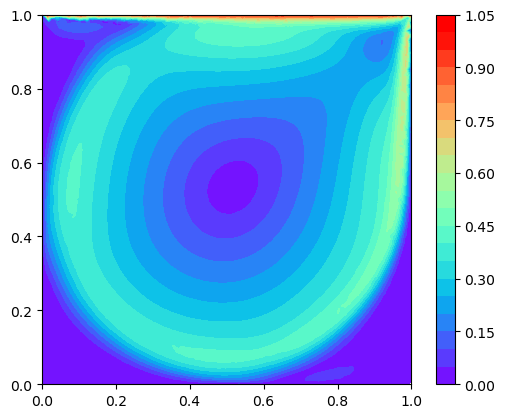} 
            \caption{}
            \label{fig3.6a}
            \end{subfigure}
        \begin{subfigure}{0.48\textwidth}
            \includegraphics[width=1\linewidth, height=6cm]
            {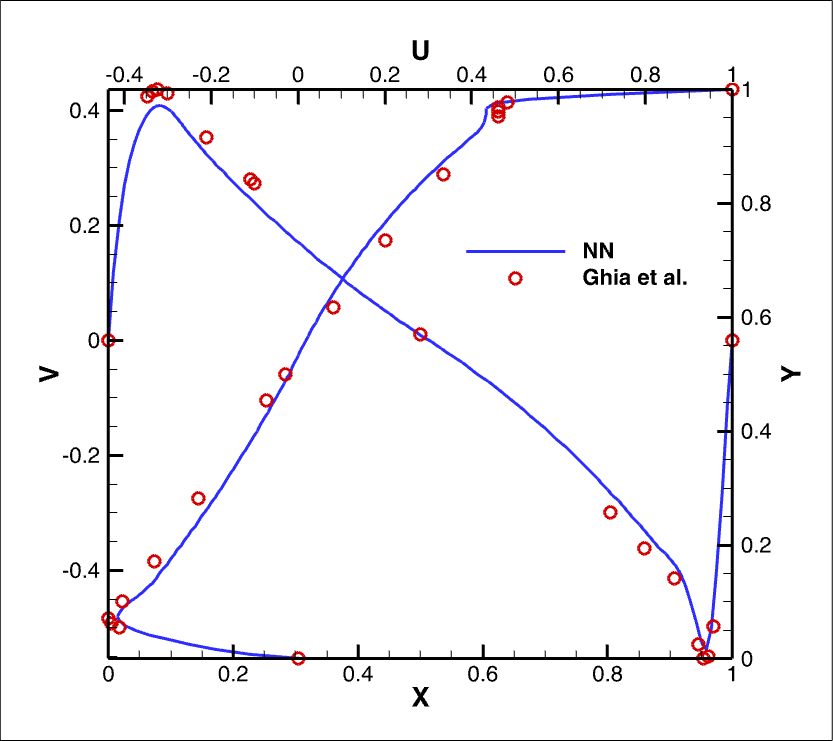}
            \caption{}
            \label{fig3.6b}
        \end{subfigure}
    \caption{(a)Re=5000 Model training result\newline(b)Re=5000 test result(Blue solid line), and Ghia et al solution(Red circle) after 
    curriculum training for total around 1,207,000 epochs}
    \label{fig3.6}
    \end{figure}

However, achieving an MAE accuracy under \(5\times10^{-1}\) has proven significantly challenging in our model for  $Re>1000$ even for the larger neural network. To alleviate the training difficulty, we implemented a curriculum training approach for improving accuracy at higher $Re$ values. In this study, we conducted predictions for $Re = 5000$ using a random V-cycle structure with denser sample points, as detailed in Table. \ref{table3.7} and \ref{table3.9}. Additional training settings are provided in the Appendix(sec.\ref{App.3}). With the current approach,
our model achieved a prediction accuracy of approximately $7\times 10^{-2}$(Fig.\ref{fig3.6}).\par
\begin{table}[htbp!]
        \centering
        \resizebox{.8\textwidth}{!}{%
        \begin{tabular}{p{3cm} | p{2cm} p{2cm} p{2cm} p{2cm} p{2cm}}
        Training sequence & 1 & 2 & 3 & 4 & 5\\
        \hline
        Reynolds number & 400 & 1000 & 2000 & 3200 & 5000\\
        \hline
        \end{tabular}%
        }
        \caption{Curriculum training sequence in Lid-driven cavity}
        \raggedright
        \label{table3.7}
\end{table}

\begin{table}[htbp!]
        \centering
        \resizebox{.6\textwidth}{!}{%
        \begin{tabular}{p{5cm} p{5cm}}
        Dataset & Collocations\\
        \hline
        \(D_1\)(main) &  14645 \\
        \(D_2\) & 3725 \\ 
        \(D_3\) & 965 \\
        \hline
        \end{tabular}%
        }
        \caption{Datasets for $Re$ = 5000}
        \raggedright
        \label{table3.9}
    \end{table}

Since the lid-driven cavity case is a well-documented benchmark in the literature, we further compare our results with selected papers from various perspectives to ensure our model achieves acceptable accuracy.\par
\begin{table}[htbp!]
        \centering
        \resizebox{1\textwidth}{!}{%
        \begin{tabular}{p{3cm} | p{3.5cm} | p{2cm} p{1cm} p{2cm} p{2cm}p{2cm} p{2cm}p{2cm} p{4cm}}
        Paper & Re cases & Method\cite{NSFnet} & Input & Training epochs & Collocation points & Accuracy for highest Re\\
        \hline
        Wang et al.\cite{Understanding_and_mitigating_gradients} & [100] & DNS(V-V) & (x,y) & unknown & unknown & 3.42e-02 \\
        Li et al\cite{li2022dynamic} & [100] & DNS(V-P) & (x,y) & 32,000 & 13,000 & 6.70e-02\\ 
        Karniadakis et al.\cite{karniadakis2023solution} & [2000,3000,5000] & DNS(V-P) & (x,y) & 3,000,000 & 120,000 & 7.00e-02\\
        Jiang et al.\cite{jiang2023applications} & [100,1000,5000,10000]\newline(1-100 samples) & DNS(V-P)\newline \underline{\textbf{FD}} & (x,y) & 200,000 & 40401 & 9.30e-03\newline(Re=1000,\newline50 samples)\\
        Wang et al.\cite{Experts_Guide} & [100,400,1000,3200] & DNS(V-P) & (x,y) & 700,000 & 8192 & 1.58e-01 \\
        Chen et al.\cite{chen2021improved} & [100,300,600] & DNS(V-V) & (x,y) & 40,000 & 1280 & 7.25e-02\\
        Chiu et al.\cite{CAN-PINN} & [400] & DNS(V-P) & (x,y) & 20,000 & 40401 & 3.16e-03 \\
        Proposed model & [400, 1000, 2000,\newline3200, 5000] & DNS(V-V) & (x,y) &1,207,000 & 19335 & 7.0e-02\\
        \hline
        \end{tabular}%
        }
        \newline
        \newline
        In method we use abbreviations provided in Jin et al.\cite{NSFnet}, where \textbf{V-P} stands for velocity-pressure formulation and \textbf{V-V} stands for vorticity-velocity formulation. As \underline{\textbf{FD}} shown in Jiang et al.\cite{jiang2023applications} means using finite difference to calculate gradients instead of automatic differentiation. As for accuracy column, the accuracy is calculated in Relative L2.
        \caption{Test detail in Lid-driven cavity}
        \raggedright
        \label{table3.8}
\end{table}
As shown in Table \ref{table3.8}, achieving high accuracy in PINNs for high Reynolds numbers (Re > 1000) typically requires extensive training epochs and a large set of collocation points unless ground truth data points are included in training loss. 
Comparing the results obtained by Karniadakis et al.\cite{karniadakis2023solution},
our model achieved an error of 7e-02, which is slightly better than the accuracy reported by Karniadakis et al. \cite{karniadakis2023solution}, while requiring significantly fewer epochs and collocation points. It is noted that as the Reynolds number increases, the rate of error reduction gradually decreases. Additionally, accurately predicting higher Reynolds numbers becomes more challenging due to the increased numerical stiffness of the governing equations, making precise adjustments to the learning rate increasingly critical.
It is also noted that although training epochs may not directly reflect time costs, this comparison indicates that our method requires fewer parameter updates and achieves higher accuracy compared to the referenced paper. Importantly, our proposed method can complement theirs without interference in implementations.\par

\section{Conclusions}\label{conclusion}
In this paper, we have demonstrated the effectiveness of Physics-Informed Neural Networks (PINNs) with our proposed method, utilizing a multi-dataset training strategy and dimensional analysis weighting. We have successfully shown that employing multiple datasets, which including at least two sizes of dataset for transfer learning, enhances the neural network's capability to fit higher frequency functions more accurately. This approach aligns with similar approach validated in MscaleDNNs\cite{MDNN,multilevel}. Additionally, we have highlighted the model's compatibility with curriculum training\cite{Failure_modes}, comparing favorably against conventional models in solving high Reynolds number steady Lid-driven cavity problems.\par

While the current work is focus on fluid dynamic problem, the current method is generic and neural network independent. Our future investigations will focus on tackling more complex problems involving higher-dimensional domains and transient tasks. Furthermore, we are keen to conduct numerical analyses on training frequencies and investigate on the effectiveness within different collocation size relationship, which potentially further optimize the performance of our method.\par

\begin{acknowledgments}
Chao-An Lin would like to acknowledge support by Taiwan National Science and Technology Council under project No. NSTC 113-2221-E-007-129.
\end{acknowledgments}

\section*{Data Availability Statement}
The data supporting this study's findings are available from the corresponding author upon reasonable request.

\section*{Conflict of Interest}
The authors have no conflicts to disclose.

\nocite{*}
\bibliography{references}
\clearpage
\section{Appendix: Detail setting of experiments}\label{App}
\subsection{1D high frequency ODE}\label{App.1}
 \begin{table}[htbp!]
        \centering
        \resizebox{.8\textwidth}{!}{%
        \begin{tabular}{p{5cm} | p{5cm}| p{5cm} }
        Dataset & $D_1$ & $D_2$  \\
        \hline
        Collocation points & 64 & 32\\
        \hline
        Optimizer & Adam(lr=1e-03) & Adam(lr=1e-03) \\
        \hline
        Scheduler \newline(ReduceLROnPlateau)&  factor=0.65, patience=100, min\_lr=1e-06 & factor=0.8, patience=100, min\_lr=1e-06 \\
        \hline
        Loss weighting \newline ($\lambda_{DE}$:$\lambda_{BC}$) & 1:1 &  1:1 \\
        \hline
        \end{tabular}%
        }
        \caption{1D high frequency ODE detail setup of fig.\ref{fig3.1}}
        \raggedright
        \label{table5.1}
    \end{table}
    
    \begin{figure}[htbp!]
    \captionsetup{justification=centering}
        \begin{subfigure}{0.48\textwidth}
            \includegraphics[width=1\linewidth, height=6cm]
            {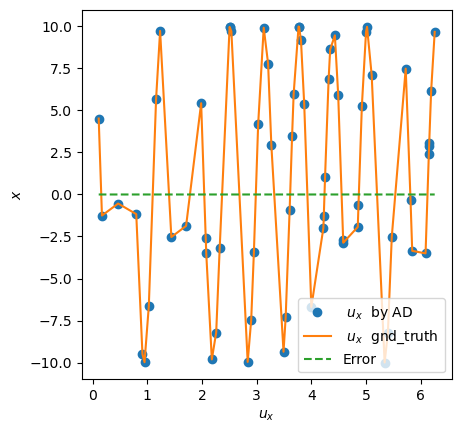} 
            \caption{PINN w/o proposed method}
            \label{fig5.1a}
            \end{subfigure}
        \begin{subfigure}{0.48\textwidth}
            \includegraphics[width=1\linewidth, height=6cm]
            {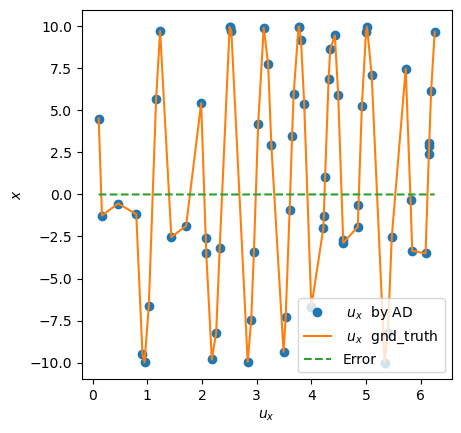}
            \caption{PINN w/ proposed method}
            \label{fig5.1b}
        \end{subfigure}
    \caption{Gradients value graphs after training for 30,000 epochs}
    \label{fig5.1}
    \end{figure}

    \begin{figure}[htbp!]
    \captionsetup{justification=centering}
        \begin{subfigure}{0.48\textwidth}
            \includegraphics[width=1\linewidth, height=6cm]
            {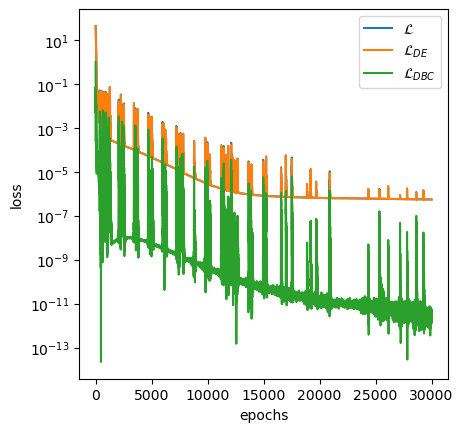} 
            \caption{PINN w/o proposed method}
            \label{fig5.2a}
            \end{subfigure}
        \begin{subfigure}{0.48\textwidth}
            \includegraphics[width=1\linewidth, height=6cm]
            {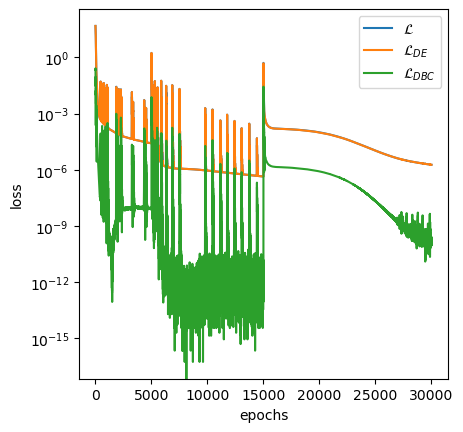}
            \caption{PINN w/ proposed method}
            \label{fig5.2b}
        \end{subfigure}
    \caption{Loss history graphs after training for 30,000 epochs}
    \label{fig5.2}
    \end{figure}
    \clearpage

\subsection{2D Smith-Hutton problem}\label{App.2}
 \begin{table}[htbp!]
        \centering
        \resizebox{.8\textwidth}{!}{%
        \begin{tabular}{p{5cm} | p{5cm}| p{5cm} |p{5cm}}
        Dataset & $D_1$ & $D_2$ & $D_3$ \\
        \hline
        Collocation points & 6670 & 1770 & 435\\
        \hline
        Optimizer & Adam(lr=5e-03) & Adam(lr=5e-03) & Adam(lr=5e-03)\\
        \hline
        Scheduler \newline(ReduceLROnPlateau) &  factor=0.8, patience=1000, min\_lr=1e-10 &  factor=0.8, patience=1000, min\_lr=1e-10 & factor=0.8, patience=1000, min\_lr=1e-10 \\
        \hline
        Loss weighting \newline ($\lambda_{DE}$:$\lambda_{NBC}$:$\lambda_{DBC}$) & 1:1:1 &  1:1:1 & 1:1:1 \\
        \hline
        \end{tabular}%
        }
        \caption{2D Smith-Hutton problem detail setup of fig.\ref{fig:sh}}
        \raggedright
        \label{table:sh}
    \end{table}

    \begin{figure}[htbp!]
    \captionsetup{justification=centering}
        \begin{subfigure}{0.4\textwidth}
            \includegraphics[width=1\linewidth, height=6cm]{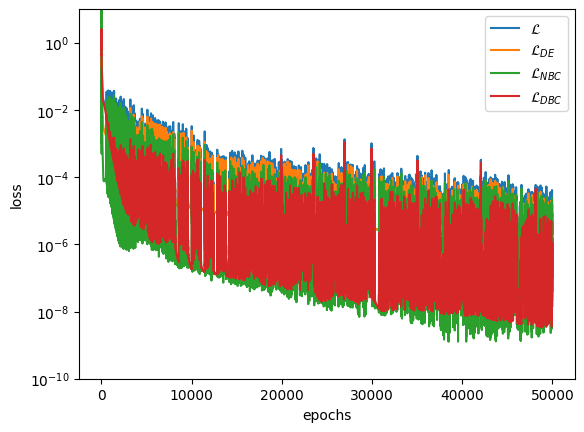} 
            \caption{uniform}
            \label{figA2.1a}
            \end{subfigure}
        \begin{subfigure}{0.4\textwidth}
            \includegraphics[width=1\linewidth, height=6cm]{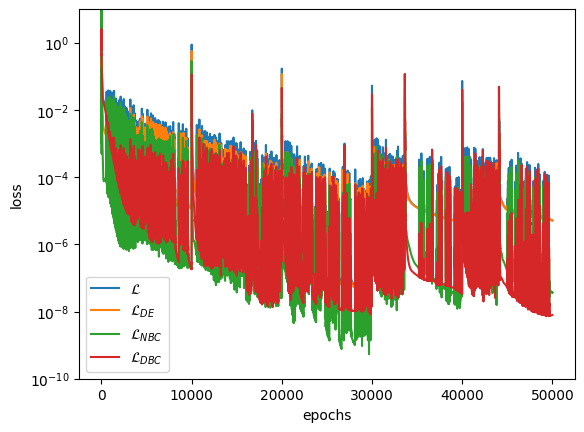}
            \caption{V cycle}
            \label{figA2.1b}
        \end{subfigure}
        \caption{plot of training history for 2D Smith-Hutton problem with different training settings.}
    \label{fig:sh-loss}
    \end{figure}

\subsection{2D steady Lid-driven cavity}\label{App.3}

\begin{table}[htbp!]
        \centering
        \resizebox{1\textwidth}{!}{%
        \begin{tabular}{p{5cm} | p{5cm}| p{5cm} | p{5cm}}
        Dataset & $D_1$ & $D_2$ & $D_3$ \\
        \hline
        Collocation points & 6565 & 1685 & 445\\
        \hline
        Optimizer & Adam(lr=1e-03) & Adam(lr=1e-03) & Adam(lr=1e-03) \\
        \hline
        Scheduler \newline(StepLR) &  factor=0.89, step\_size=1000 & factor=0.93, step\_size=1000 &  factor=0.97, step\_size=1000 \\
        \hline
        Loss weighting \newline ($\lambda_{DE}$:$\lambda_{BC}$) & $DW_{root}$(eq.\ref{eq11}) &  $DW_{root}$ & $DW_{root}$\\
        \hline
        \end{tabular}%
        }
        \caption{2D Lid-driven cavity detail setup of Re = 400(fig.\ref{fig3.4})}
        \raggedright
        \label{table5.2}
\end{table}

\begin{figure}[htbp!]
    \captionsetup{justification=centering}
        \begin{subfigure}{0.48\textwidth}
            \includegraphics[width=1\linewidth, height=6cm]
            {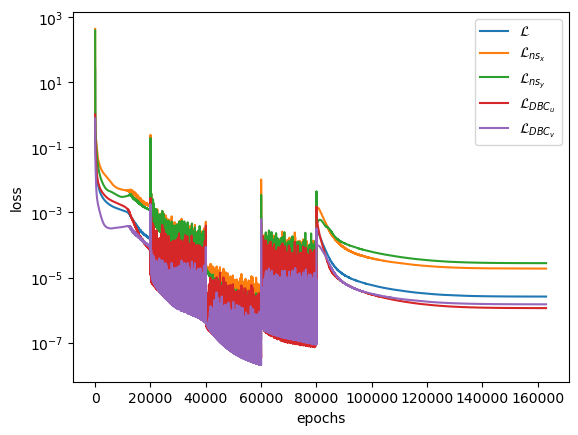} 
            \caption{}
            \label{fig5.3a}
            \end{subfigure}
        \begin{subfigure}{0.48\textwidth}
            \includegraphics[width=1\linewidth, height=6cm]
            {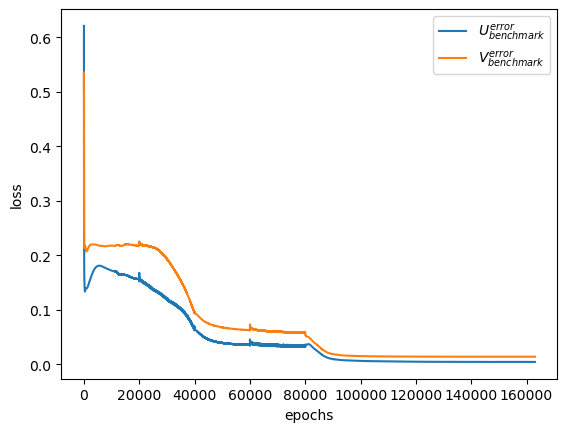}
            \caption{}
            \label{fig5.3b}
        \end{subfigure}
    \caption{(a)Loss history and (b)Error history for Re=400 (fig.\ref{fig3.4})}
    \label{fig5.3}
    \end{figure}

\begin{table}[htbp!]
        \centering
        \resizebox{1\textwidth}{!}{%
        \begin{tabular}{p{5cm} | p{5cm}| p{5cm} | p{5cm}}
        Dataset & $D_1$ & $D_2$ & $D_3$ \\
        \hline
        Collocation points & 6565 & 1685 & 445\\
        \hline
        Optimizer & Adam(lr=1e-03) & Adam(lr=1e-03) & Adam(lr=1e-03) \\
        \hline
        Scheduler \newline(ReduceLROnPlateau) &  factor=0.78, patience=1000, min\_lr=1e-08 & factor=0.82, patience=1000, min\_lr=1e-08 &  factor=0.90, patience=1000, min\_lr=1e-08 \\
        \hline
        Loss weighting \newline ($\lambda_{DE}$:$\lambda_{BC}$) & $DW_{root}$(eq.\ref{eq11}) &  $DW_{root}$ & $DW_{root}$\\
        \hline
        \end{tabular}%
        }
        \caption{2D Lid-driven cavity detail setup of Re = 1000(fig.\ref{fig3.5})}
        \raggedright
        \label{table5.3}
\end{table}

\begin{figure}[htbp!]
    \captionsetup{justification=centering}
        \begin{subfigure}{0.48\textwidth}
            \includegraphics[width=1\linewidth, height=6cm]{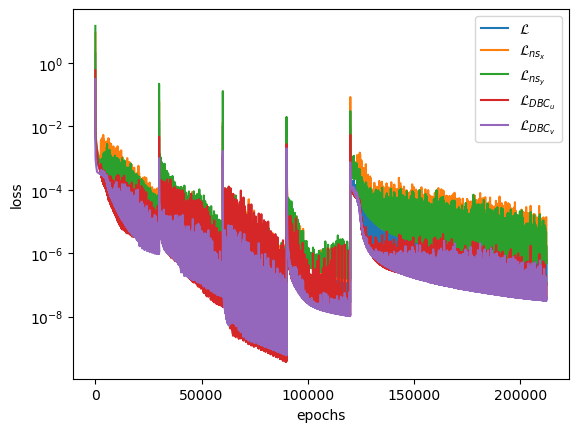} 
            \caption{}
            \label{fig5.4a}
            \end{subfigure}
        \begin{subfigure}{0.48\textwidth}
            \includegraphics[width=1\linewidth, height=6cm]{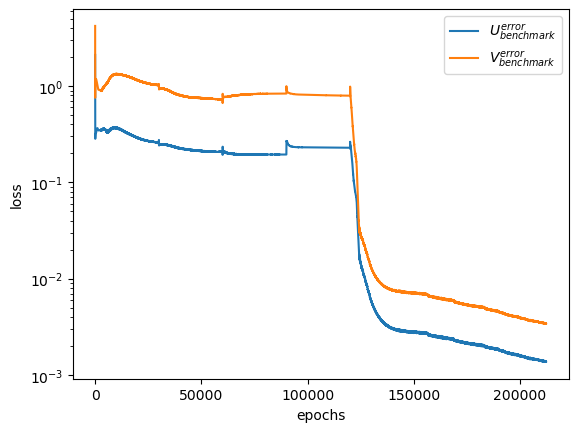}
            \caption{}
            \label{fig5.4b}
        \end{subfigure}
    \caption{(a)Loss history and (b)Error history for Re=1000 (fig.\ref{fig3.5})}
    \label{fig5.4}
    \end{figure}

    \begin{table}[htbp!]
        \centering
        \resizebox{1\textwidth}{!}{%
        \begin{tabular}{p{5cm} | p{3cm}| p{3cm} | p{3cm} | p{3cm} | p{3cm}}
        Reynolds number & 400 & 1000 & 2000 & 3200 & 5000 \\
        \hline
        \hline
        Level Duration & 20,000 & 20,000 & 20,000 & 30,000 & 30,000\\
        \hline
        Initial LR & 1e-3 & 1e-3 & 1e-4 & 1e-4 & 4e-5/4e-5/1e-4 \\
        \hline
        Scheduler factor\newline(ReduceLROnPlateau) & 0.73/0.80/0.88  &  
        0.68/0.75/0.83 & 0.63/0.70/0.81 & 0.58/0.65/0.79 & 0.58/0.6/0.80 \\
        \hline
        Loss weighting \newline ($\lambda_{DE}$:$\lambda_{BC}$) & $DW_{root}$(eq.\ref{eq11}) &  $DW_{root}$ & $DW_{root}$ & $DW_{root}$ & $DW_{root}$\\
        \hline
        Total loss criterion & 1e-07 & 5e-08 & 5e-07 & 5e-08 & 1e-08\\
        \hline
        \hline
        Error value (u/v)& 1.30e-02/1.43e-01 &  1.36e-02/2.3e-02 & no reference values & 5.75e-02/6.69e-02 & 5.86e-02/6.45e-02\\ 
        \hline
        Training epochs & 111,297 & 143,225 & 82,016 & 370,844 & 500,000\\
        \hline
        Total epochs & 111,297 & 254,522 & 336,538 & 707,382 & 1,207,382\\
        \hline
        \end{tabular}%
        }
        \newline
        \\
        The setting is similar to fig.\ref{fig3.6} and Table.\ref{table5.3}
        \caption{2D Lid-driven cavity curriculum detail setup of Re = 5000 (Table. \ref{table3.9})}
        \raggedright
        \label{table5.5}
\end{table}
    
\end{document}